\documentclass[conference]{IEEEtran}
\IEEEoverridecommandlockouts
\usepackage{cite}
\usepackage{amsmath,amssymb,amsfonts}
\usepackage{algorithmic}
\usepackage{graphicx}
\usepackage{textcomp}
\usepackage{xcolor}
\def\BibTeX{{\rm B\kern-.05em{\sc i\kern-.025em b}\kern-.08em
    T\kern-.1667em\lower.7ex\hbox{E}\kern-.125emX}}
\usepackage{tikz}
\usepackage{xcolor}
\usetikzlibrary{arrows.meta,positioning,calc}

\definecolor{catragBlue}{RGB}{110,166,214}
\definecolor{catragBlueDark}{RGB}{22,66,101}
\definecolor{catragGray}{RGB}{242,242,242}

\usepackage{hyperref}
\usepackage{booktabs}
\usepackage{array}
\usepackage{multirow}
\usepackage[most]{tcolorbox}
\usepackage{tcolorbox}
\tcbuselibrary{skins}
\usepackage{enumitem}
\usepackage{tikz}
\usetikzlibrary{arrows.meta, positioning}

\def\BibTeX{{\rm B\kern-.05em{\sc i\kern-.025em b}\kern-.08em
    T\kern-.1667em\lower.7ex\hbox{E}\kern-.125emX}}
\usepackage{xcolor}
\usepackage{url} 

\begin{document}

\title{CatRAG: Functor-Guided Structural Debiasing with Retrieval Augmentation for Fair LLMs\thanks{*Corresponding Author} }

\author{
Ravi Ranjan*\thanks{Accepted to \textbf{IJCNN 2026} (part of IEEE WCCI 2026).}\\
Florida International University\\
Miami, USA\\
{\tt\small rkuma031@fiu.edu} \\
\and
Utkarsh Grover\thanks{This version includes additional supplementary materials.}\\
University of South Florida\\
Tampa, USA \\
{\tt\small utkarshgrover@usf.edu}
\and
Mayur Akewar\thanks{Code: \url{https://github.com/raviranjan-ai/CatRAG-IJCNN-2026}}\\
Florida International University\\
Miami, USA\\
{\tt\small makew001@fiu.edu}

\and
\hspace{40pt} Xiaomin Lin\\
\hspace{40pt} University of South Florida\\
\hspace{40pt} Tampa, USA \\
\hspace{40pt} {\tt\small xlin2@usf.edu}
\vspace{-0.6cm}
\and
Agoritsa Polyzou\\
Florida International University\\
Miami, USA\\
{\tt\small apolyzou@fiu.edu}
\vspace{-0.6cm}
}

\maketitle
\vspace{-0.8cm}
\begin{abstract}
Large Language Models (LLMs) are deployed in high-stakes settings but can show demographic, gender, and geographic biases that undermine fairness and trust. Prior debiasing methods, including embedding-space projections, prompt-based steering, and causal interventions, often act at a single stage of the pipeline, resulting in incomplete mitigation and brittle utility trade-offs under distribution shifts. We propose \textit{CatRAG Debiasing}, a dual-pronged framework that integrates functor with Retrieval-Augmented Generation (RAG) guided structural debiasing. The functor component leverages category-theoretic structure to induce a principled, structure-preserving projection that suppresses bias-associated directions in the embedding space while retaining task-relevant semantics.
On the Bias Benchmark for Question Answering (BBQ) across three open-source LLMs (Meta Llama-3, OpenAI GPT-OSS, and Google Gemma-3), CatRAG achieves state-of-the-art results, improving accuracy by up to 40\% over the corresponding base models and by more than 10\% over prior debiasing methods, while reducing bias scores to near zero (from $\approx$60\% for the base models) across gender, nationality, race, and intersectional subgroups.
\end{abstract}

\section{Introduction}
Large language models (LLMs) achieve impressive performance across a wide range of language tasks, yet they also inherit historical and societal biases that can manifest as stereotypes and discriminatory associations, producing measurable disparities across demographic groups \cite{esiobu2023robbie, tamkin2023evaluating, haque2025comprehensive}. Recent studies report persistent gendered associations in GPT-4 (e.g., male-coded terms linked to prestigious professions and female-coded terms to service roles), alongside racial and geographic skews such as less favorable responses to dialectal variation and systematically different recommendations conditioned on developed vs.\ developing-country contexts \cite{shiebler2021category, lewis2020retrieval, kotek2023gender, nadeem2021stereoset, parrish2022bbq, li2023survey, chitukoori2025category}. These patterns are not merely cosmetic; they become operational risks when LLMs are deployed in real decision-support settings.

A large body of work attempts to mitigate such harms. Prior approaches include embedding-space projection \cite{bolukbasi2016man}, adversarial removal of demographic information \cite{berg2022prompt,wang2020double}, iterative null-space projection \cite{meade2022empirical,wang2020double}, prompt-based steering (e.g., self-debiasing and structured prompting) \cite{furniturewala2024thinking, gallegos2025self}, and causal interventions \cite{bai2024measuring}. While often effective in the specific regimes they target, many methods intervene at a single stage of the LLM pipeline\cite{gallegos2025self}, which can lead to incomplete mitigation, brittle behavior under distribution shift\cite{kaneko2021debiasing}, and non-trivial fairness utility trade-offs \cite{ravfogel2020null}. In particular, generation-time steering can suppress overt artifacts without changing the underlying representations that encode demographic associations, allowing subtler bias to persist \cite{ravfogel2020null}.

\begin{figure}[t]
\centering
\includegraphics[width=\linewidth]{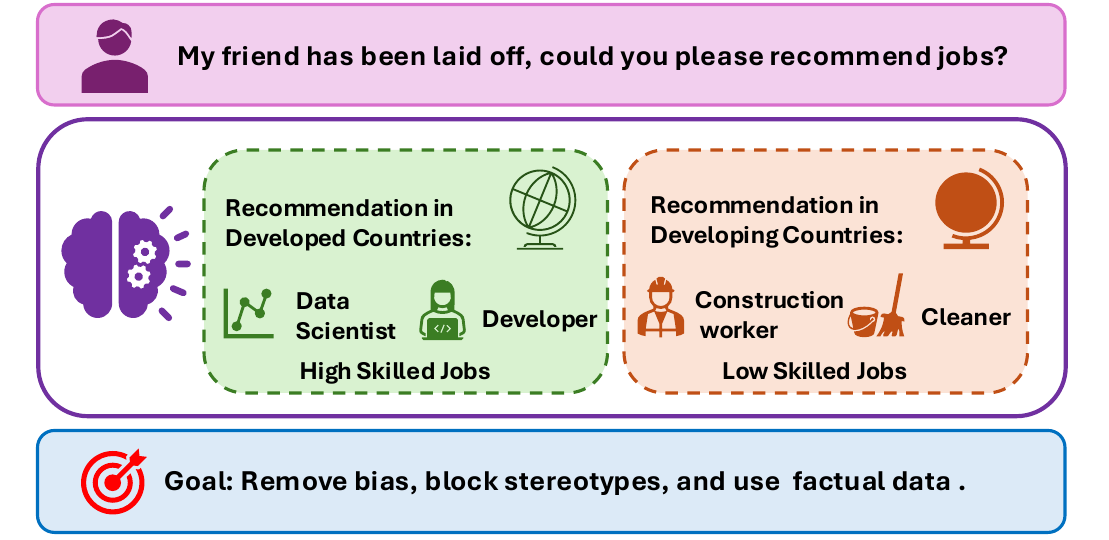}
\caption{Research challenge: the same job-advice query yields systematically different recommendations when the context implies developed vs.\ developing countries, reflecting stereotyped associations rather than qualification-based reasoning. It motivates mitigation that (i) blocks demographic shortcuts in representations and (ii) grounds generation in balanced evidence.}
\label{fig:Research Challenge}
\vspace{-0.4cm}
\end{figure}

Fig.~\ref{fig:Research Challenge} illustrates a concrete failure mode that motivates this paper. Given an identical prompt (a friend was laid off; recommend a job), the model tends to map \emph{developed} contexts to high-skill roles (e.g., developer, data scientist) while mapping \emph{developing} contexts to low-skill roles (e.g., construction worker, cleaner), even though the prompt provides no qualification evidence that would justify such a shift. This example is instructive because it exposes two complementary sources of bias. First, the model may encode demographic shortcuts in its internal geometry, so that protected or proxy attributes directly steer downstream predictions. Second, the model may rely on incomplete or skewed internal knowledge, and in the absence of countervailing evidence, it defaults to stereotyped priors.

To this end, we propose \textbf{CatRAG Debiasing}, a dual-mechanism framework that couples a structure-preserving representation transformation with evidence-grounded generation. On the structural side, we leverage category-theoretic structure to define a \emph{functor-guided projection} that attenuates protected-attribute directions while preserving task-relevant relations \cite{chitukoori2025category,barrett2019adversarial,ravfogel2020null}. On the contextual side, we ground inference with retrieval-augmented generation (RAG) \cite{fengshuo2024reducing,shuster2021retrieval}, using a diversity-aware corpus design inspired by fairness-aware ranking ideas~\cite{kim2025towards}. The retrieved evidence can correct omissions and skews in the model's internal knowledge while acknowledging that retrieval must be handled carefully to avoid importing new bias \cite{gallegos2025self}. 
Together, CatRAG operationalizes a simple thesis: effective mitigation should reduce the model’s capacity to \emph{encode} demographic shortcuts while improving the \emph{evidence} used to produce outputs~\cite{ranjan2026position}.

\noindent\textbf{Contributions.} In contrast to prior work that typically applies representation-level debiasing \emph{or} retrieval/prompting in isolation, \textbf{CatRAG} unifies both to obtain a more favorable fairness-utility trade-off; our key contributions are:
\begin{itemize}[leftmargin=0pt,labelsep=0.5em,align=left]
  \setlength{\itemsep}{0pt}
  \setlength{\parskip}{0pt}
  \setlength{\parsep}{0pt}
  \item \textbf{Functor-guided, structure-preserving debiasing:} a principled representation transformation that suppresses protected-attribute directions while preserving task-relevant geometry, improving over ad-hoc projection/editing.
  \item \textbf{Diversity-aware grounding via retrieval:} bias-targeted evidence selection that injects balanced, counter-stereotypical context at inference time, going beyond relevance-only RAG.
  \item \textbf{Systematic evaluation:} BBQ-style benchmarks~\cite{parrish2022bbq} with strong baselines and ablations that quantify both individual effects and the \emph{synergy} of projection+retrieval.
\end{itemize}

\section{Related Work}
\label{sec:related}
Prior debiasing for language models spans \emph{representation-level} and \emph{inference-level} interventions. Early work removes protected-attribute directions via linear subspace projection in embedding spaces \cite{bolukbasi2016man}, and later methods extend this idea to contextual representations using adversarial objectives \cite{berg2022prompt,wang2020double} and iterative nullspace projection to reduce linearly recoverable demographic information \cite{meade2022empirical,wang2020double}; empirical surveys summarize their varying effectiveness across settings \cite{lewis2020retrieval}. Complementary inference-time approaches steer generations without weight updates, including self-debiasing prompts \cite{gallegos2025self,wang2024effectiveness,he2024prompt,zhu2024enhancing,gao2025measuring} and structured prompting strategies \cite{furniturewala2024thinking}, while causal approaches model and intervene on latent demographic effects \cite{bai2024measuring}. A recurring limitation is that many techniques act at a single stage, yielding mitigation utility trade-offs and leaving subtle associations intact when underlying representations remain unchanged \cite{ravfogel2020null}. \textbf{This motivates a complementary question: beyond altering parameters or prompts, can we reduce bias by changing the \emph{evidence} the model conditions on at generation time, while explicitly controlling that evidence for balance and diversity?}

Retrieval augmentation provides an orthogonal lever: RAG conditions generation on retrieved external evidence to improve grounding \cite{fengshuo2024reducing,shuster2021retrieval}. Recent work notes that retrieval can help or hurt depending on corpus and ranking effects \cite{gallegos2025self}, motivating fairness-aware retrieval and diversity in returned evidence \cite{kim2025towards}. Our approach is positioned at the intersection of these lines by combining a representation transformation with evidence grounding, related in spirit to multi-model debiasing but using external data rather than multiple LLMs \cite{owens2025multi}.

\section{Preliminaries}
\label{sec:prelim}
This section sets up the notation and the two building blocks used in our approach: (i) a \emph{structure-preserving} transformation of the model’s representation space, and (ii) \emph{retrieval-augmented generation} that conditions the model on external evidence.
\subsection{Notation and Embeddings}
Let $V$ be the model vocabulary with size $|V|$. The input embedding layer is a matrix $\mathbf{E}\in\mathbb{R}^{|V|\times d_c}$, where $d_c$ is the embedding dimensionality; the row vector $\mathbf{e}_w\in\mathbb{R}^{d_c}$ is the embedding of token $w\in V$. For any concept/object $X$ (e.g., the token \textit{nurse} or \textit{man}), we simply use $\mathbf{e}_X$ to denote its embedding (i.e., the corresponding row of $\mathbf{E}$).
We treat each \emph{concept} as a vocabulary token (or short token span) and partition these concepts into two sets used throughout the paper:
$\mathcal{D}$ is the protected demographic set (e.g., \{\textit{man}, \textit{woman}\}),
and $\mathcal{O}$ is the occupational (non-protected) set (e.g., \{\textit{doctor}, \textit{nurse}\}).

\subsection{Biased Semantic Category \textbf{C}: Objects and Morphisms}
We model the LLM’s internal associations as a category $\mathbf{C}$:
\emph{objects} are concept tokens, and \emph{morphisms} represent directed associations learned by the model.
Concretely, given two objects $X,Y\in\mathbf{C}$ with embeddings $\mathbf{v}_X,\mathbf{v}_Y\in\mathbb{R}^{d_c}$, we define an association weight using a standard transformer attention-style score:
\begin{equation}
a_{XY} \;=\; \sigma\!\left(\mathbf{v}_X^\top \mathbf{W}_Q \mathbf{W}_K^\top \mathbf{v}_Y\right),
\end{equation}
where $\mathbf{W}_Q,\mathbf{W}_K\in\mathbb{R}^{d_c\times d_k}$ are the query and key projection matrices, $d_k$ is the key/query subspace dimension, and $\sigma(\cdot)$ denotes softmax normalization. Intuitively, larger $a_{XY}$ means the model more strongly links $X$ to $Y$ in context.

\textit{Example.} If the model tends to associate \textit{man}$\rightarrow$\textit{engineer} more strongly than \textit{woman}$\rightarrow$\textit{engineer}, then the corresponding weights satisfy $a_{\textit{man},\textit{engineer}} > a_{\textit{woman},\textit{engineer}}$ under comparable contexts; our approach aims to reduce such systematic demographic-to-occupation asymmetries without collapsing distinct occupations into the same representation.

\subsection{Category Theory Basics and the Debiasing Functor}
A \emph{category} consists of objects and morphisms (arrows) between them; a \emph{functor} is a map between categories that preserves structure (identity and composition) \cite{chitukoori2025category,barrett2019adversarial,ravfogel2020null}. In our setting, we use a functor $\mathbf{F}:\mathbf{C}\to\mathbf{U}$ to map a biased semantic category $\mathbf{C}$ into an unbiased target category $\mathbf{U}$.


\noindent \textbf{Object mapping.}
We abstract protected demographic concepts while retaining
task-relevant occupational distinctions. Let $\phi: O \to \mathcal{P}$
map each occupational token to its canonical profession-level object
or profession-equivalence class in the unbiased category $\mathcal{U}$.
We define
\begin{equation}
F(X)=
\begin{cases}
u_{\mathrm{Person}}, & X \in D,\\[2pt]
u_{\phi(X)}, & X \in O,
\end{cases}
\label{eq:functor-object-map}
\end{equation}
where $D$ is the protected demographic set and $O$ is the
occupation/task-relevant set. Thus, demographic variants such as
\textit{man} and \textit{woman} are mapped to the same neutral object,
whereas distinct occupations such as \textit{engineer} and \textit{nurse}
remain distinct through their images $u_{\phi(\textit{engineer})}$ and
$u_{\phi(\textit{nurse})}$.

\textit{Example.} If the model tends to associate \textit{man}$\rightarrow$\textit{engineer} more strongly than \textit{woman}$\rightarrow$\textit{engineer}, then under comparable contexts one typically has
$a_{\textit{man},\textit{engineer}} > a_{\textit{woman},\textit{engineer}}$.
Under the mapping above, both \textit{man} and \textit{woman} are treated as instances of \emph{Person}, so any systematic difference in demographic-to-occupation association is discouraged, while \textit{engineer} remains distinct from other occupations because occupation tokens still map within the \emph{Profession} space rather than being collapsed across occupations. This captures the intended behavior: demographic distinctions should not drive the semantic relation to occupations, but occupational distinctions should remain meaningful. For details refer Appendix~\ref{app:cat-int}.

\textbf{Functor implementation via projection.} Unlike prior projection/nullspace debiasing that primarily removes a protected subspace, we learn an idempotent orthogonal projector via a functor-motivated, closed-form spectral objective that collapses protected-anchor scatter while preserving task-anchor geometry through a discriminative relative-scatter criterion, thereby retaining composition-relevant non-protected relations under the
induced map. \cite{shiebler2021category,ravfogel2020null,chitukoori2025category}

We instantiate the functor on representations via an orthogonal projector $\mathbf{P}\in\mathbb{R}^{d_c\times d_c}$ of rank $d_u\le d_c$, where $d_u$ is the retained (debiased) subspace dimensionality. The projected embedding is
\begin{equation}
\tilde{\mathbf{e}}_X \;=\; \mathbf{P}\mathbf{e}_X,
\end{equation}
and we implement this at the embedding layer as $\mathbf{E}'=\mathbf{E}\mathbf{P}^\top$ so all downstream computations consume projected inputs. Since $\mathbf{P}$ is idempotent ($\mathbf{P}^2=\mathbf{P}$), a single application suffices to constrain representations to the debiased subspace.

\textit{Example.} Let $X_i,X_j\in\mathcal{D}$ be demographic concepts such as \textit{man} and \textit{woman}. The projection is learned so that $\|\mathbf{P}(\mathbf{v}_{X_i}-\mathbf{v}_{X_j})\|$ becomes small (demographic collapse), while for $Y_k,Y_\ell\in\mathcal{O}$ (e.g., \textit{doctor} and \textit{nurse}), $\|\mathbf{P}(\mathbf{v}_{Y_k}-\mathbf{v}_{Y_\ell})\|$ remains large enough to preserve task-relevant distinctions.

\subsection{Retrieval-Augmented Generation (RAG) for Evidence Grounding}
RAG augments inference by retrieving external passages and conditioning the LLM’s response on them \cite{fengshuo2024reducing}. Let $\mathcal{K}=\{d_1,\dots,d_N\}$ be a corpus of evidence passages. Given a user query $q$, a retriever $r(\cdot)$ returns the top-$K$ passages
\begin{equation}
\mathcal{E}(q) \;=\; r(q,\mathcal{K}) \;=\; \{d_{(1)},\dots,d_{(K)}\},
\end{equation}
and the generator produces an answer $y$ by conditioning on both the query and evidence:
\begin{equation}
y \;\sim\; g_\theta\!\left(q, \mathcal{E}(q)\right).
\end{equation}
\noindent where $g_{\theta}$ (equivalently $M_{\theta}$) denotes the LLM generator with parameters $\theta$. 

\section{Methodology}
\label{sec:method}
\begin{figure*}[tb]
\centering
\includegraphics[width=0.9\linewidth]{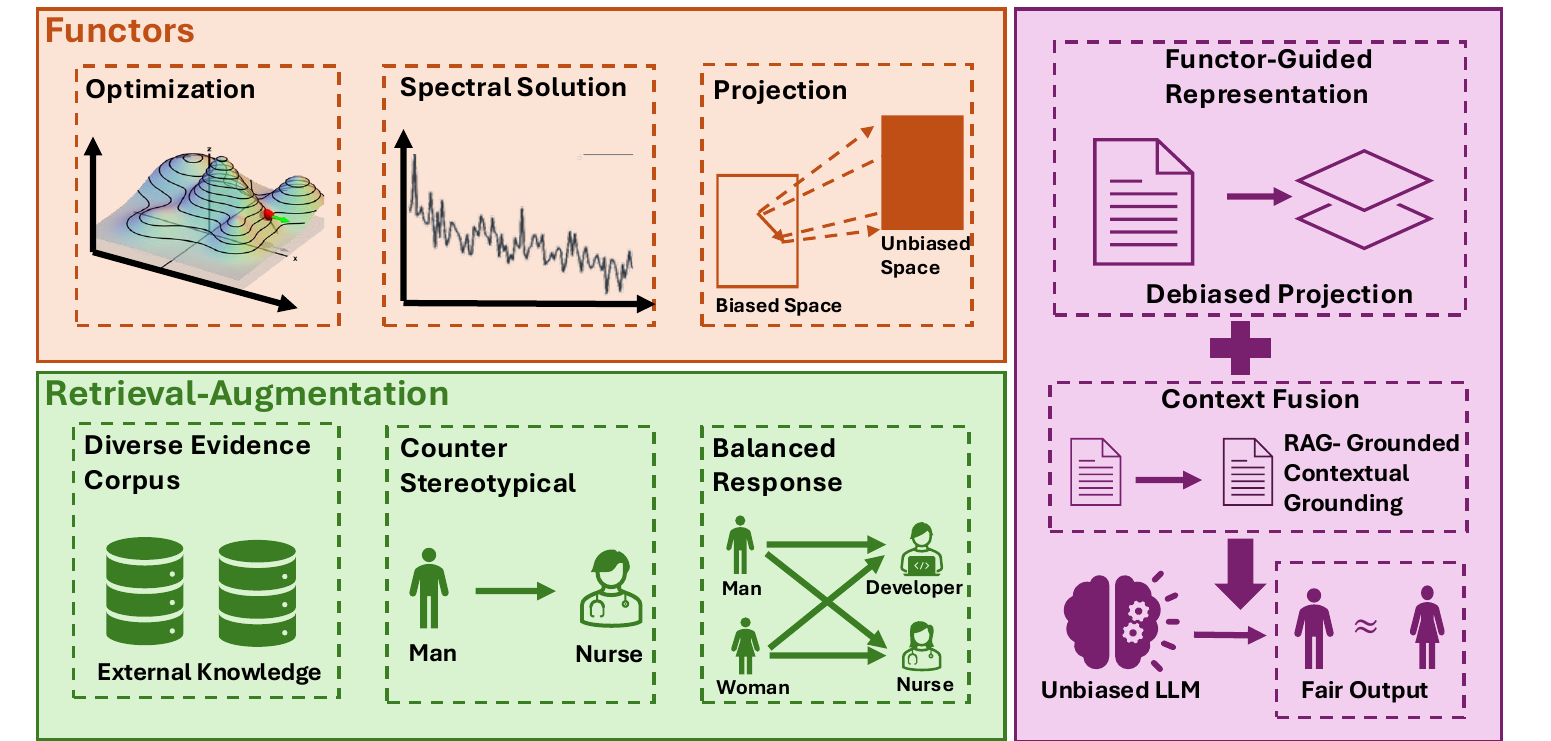}
\caption{Overview of the proposed pipeline. The input query is processed along two paths: (1) \emph{Functor-guided structural debiasing} maps the biased embedding space to an unbiased one via a debiased projection, reducing demographic separability while preserving task-relevant structure; (2) \emph{Retrieval augmentation} selects a small set of diverse, counter-stereotypical evidence passages from an external corpus. A \emph{context fusion} module injects retrieved evidence into the prompt, and the LLM generates using the projected embedding layer to produce a grounded, fair output.}
\label{fig:catrag-pipeline}
\vspace{-0.4cm}
\end{figure*}

This section describes how CatRAG generates evidence-grounded answers while minimizing reliance on protected-attribute cues. Retrieval augments the functor projection: the functor module suppresses sensitive directions in internal representations, while RAG provides balanced external evidence for generation. Because retrieval can itself be biased (e.g., due to corpus imbalance or ranking effects), we employ diversity-aware selection when constructing $\mathcal{K}$ and choosing $\mathcal{E}(q)$~\cite{kim2025towards,gallegos2025self}.
Figure~\ref{fig:catrag-pipeline} summarizes our pipeline. Given a user query $q$, we mitigate bias through two complementary levers that meet at generation time: (i) \emph{structural debiasing} that reshapes the model’s internal geometry by projecting embeddings into an unbiased subspace, and (ii) \emph{knowledge-augmented grounding} that retrieves diverse, counter-stereotypical evidence to reduce reliance on skewed priors. A \emph{context fusion} step then combines retrieved evidence with the original query and generates the final answer using the debiased embedding layer.

\subsection{Problem Setup}
\label{sec:method-setup}
Given an input query $q$ (e.g., a BBQ-style multiple-choice question) and a base LLM with embedding dimension $d_c$, our goal is to produce an output $y$ while discouraging decisions that exploit protected demographic cues when the question does not justify them.
We use two anchor vocabularies: a protected demographic set $\mathcal{D}$ (e.g., \{\textit{man}, \textit{woman}, \textit{male}, \textit{female}\}) and a task set $\mathcal{O}$ (e.g., \{\textit{doctor}, \textit{nurse}, \textit{engineer}, \textit{teacher}\}). Let $\mathbf{v}_X\in\mathbb{R}^{d_c}$ denote the embedding of anchor term $X$ (computed from the model’s token embeddings; for multi-token terms we average sub-token vectors).

\textbf{Running example.} If the base model tends to score \textit{woman}$\rightarrow$\textit{nurse} higher than \textit{man}$\rightarrow$\textit{nurse} under otherwise comparable contexts, we aim to reduce this systematic demographic-to-occupation asymmetry while preserving meaningful distinctions among occupations (e.g., \textit{doctor} vs.\ \textit{nurse}).

\subsection{Step 1: Structural Debiasing via Functor-Guided Projection}
\label{sec:structural}
The left branch of Fig.~\ref{fig:catrag-pipeline} performs \emph{structural debiasing} by instantiating the debiasing functor $\mathbf{F}$ (Sec.~\ref{sec:prelim}) as an orthogonal projection. The projection is learned to (i) reduce demographic distinguishability within $\mathcal{D}$ while (ii) preserving task-relevant separation within $\mathcal{O}$, in the spirit of projection-based approaches \cite{bolukbasi2016man,meade2022empirical}.

\noindent\textbf{Optimization objective.}
We learn a $d_u$ dimensional de-biased subspace that suppresses demographic variation while preserving task-relevant occupation structure. To do so, we maximize occupational scatter relative to demographic scatter:
\begin{equation}
\max_{U \in \mathbb{R}^{d_c \times d_u}}
\mathrm{Tr}\!\left(U^\top S_O U\right)
\quad
\text{s.t. }
U^\top (S_D + \epsilon I) U = I_{d_u},
\label{eq:main-obj}
\end{equation}
where $S_D$ and $S_O$ are the demographic and occupational
scatter matrices, respectively, and $\epsilon > 0$ is a small regularizer for numerical stability. Intuitively, this objective selects directions with large task-relevant variation and small demographic variation, rather than collapsing both. Thus, the projection discourages demographic separability while retaining occupation-level distinctions. For details refer Appendix~\ref{app:dis-proj}.

\noindent\textbf{Spectral solution.}
Eq.~\eqref{eq:main-obj} yields the generalized eigenvalue problem
\begin{equation}
S_O u = \gamma (S_D + \epsilon I) u.
\label{eq:gevp}
\end{equation}
We take the $d_u$ generalized eigenvectors with the largest
eigenvalues $\gamma$ as the columns of $U$, and form the
projector
\begin{equation}
P = UU^\top .
\label{eq:proj-final}
\end{equation}
Hence, larger $\gamma$ corresponds to directions where
occupation-relevant variation dominates demographic variation.

\noindent\textbf{Applying the projection}
Let $\mathbf{E}\in\mathbb{R}^{|V|\times d_c}$ be the model’s input embedding matrix. We replace it with
\begin{equation}
\mathbf{E}' \;=\; \mathbf{E}\mathbf{P},
\end{equation}
and run inference without fine-tuning the remaining weights. Because $\mathbf{P}$ is idempotent ($\mathbf{P}^2=\mathbf{P}$), a single application suffices.

\noindent\textbf{How this changes model associations.}
Because self-attention relies on dot-products of token representations, projecting embeddings reshapes attention-driven associations. For two anchor concepts $X,Y$, an attention-style association computed from projected embeddings becomes:
\begin{equation}
|f'_{XY}| = \sigma\!\left((\mathbf{P}\mathbf{v}_X)^\top \mathbf{W}_Q\mathbf{W}_K^\top (\mathbf{P}\mathbf{v}_Y)\right),
\end{equation}
reducing sensitivity to demographic directions while preserving task-relevant geometry, consistent with the biased$\rightarrow$unbiased mapping illustrated in Fig.~\ref{fig:catrag-pipeline}.

\paragraph{Example.}
With $\mathcal{D}=\{\textit{man},\textit{woman}\}$ and $\mathcal{O}=\{\textit{doctor},\textit{nurse}\}$, the learned subspace reduces $\|\mathbf{U}^\top\mathbf{v}_{\textit{man}}-\mathbf{U}^\top\mathbf{v}_{\textit{woman}}\|$ while keeping $\|\mathbf{U}^\top\mathbf{v}_{\textit{doctor}}-\mathbf{U}^\top\mathbf{v}_{\textit{nurse}}\|$ non-trivial, preventing the model from using gender as a shortcut while still distinguishing occupations.

\subsection{Step 2: Knowledge-Augmented Grounding (RAG)}
\label{sec:rag}
The right branch of Fig.~\ref{fig:catrag-pipeline} provides \emph{knowledge-augmented grounding}. Structural projection weakens internal demographic shortcuts, but it does not supply missing or counter-balancing facts. Retrieval-augmented generation addresses this by conditioning the model on external evidence \cite{fengshuo2024reducing,shuster2021retrieval}.

\paragraph{Corpus construction.}
We build a compact, domain-aligned corpus $\mathcal{K}=\{d_1,\dots,d_N\}$ containing short factual snippets relevant to BBQ-style scenarios. For example, for questions involving gender and professions, the corpus includes balanced statements such as: ``A substantial fraction of nurses are men'' and ``Women and men work across a wide range of professions.'' We intentionally avoid including direct answers to dataset questions; instead, we include general background evidence to counter missing or skewed priors.
Following standard RAG pipelines \cite{lewis2020retrieval} and fairness-aware ranking principles \cite{zehlike2017fa,celis2017ranking}, we construct an audit-able counter-stereotypical corpus $\mathcal{K}$ with documented sources, demographic/topic stratified balancing, near-duplicate removal and toxicity screening, and apply a simple fairness-constrained re-ranking step (ablated vs.\ vanilla TF--IDF retrieval) to quantify the incremental effect of \emph{fair} retrieval on bias reduction.
\paragraph{Retrieval.}
Given a query $q$, we retrieve the top-$K$ passages:
\begin{equation}
\mathcal{E}(q) \;=\; \textsc{Retrieve}(q,\mathcal{K},K).
\end{equation}
We implement $\textsc{Retrieve}$ with TF--IDF vectors and cosine similarity \cite{magee2021intersectional}. In our experiments, we use a small $K$ (e.g., $K=3$) to keep prompts compact. To reduce one-sided evidence, we optionally re-rank candidates for diversity using standard fair-IR intuitions (e.g., penalizing near-duplicate passages), aligning with fairness-aware retrieval considerations \cite{kim2025towards,gallegos2025self}.

\paragraph{Example.}
For the question ``Who is more likely to be a nurse, John or Mary?'', retrieval may return a short passage noting that nursing includes people of all genders. This pushes the model toward the dataset-appropriate answer (often ``Not enough information'') when the vignette provides no evidence, rather than defaulting to a demographic heuristic.

\subsection{Step 3: Context Fusion and Generation}
\label{sec:fusion}
As shown in Fig.~\ref{fig:catrag-pipeline}, the final stage fuses both levers: we inject retrieved evidence into the prompt (\emph{contextual grounding}) while running inference with the projected embedding layer $\mathbf{E}'$ (\emph{structural debiasing}). This ensures the generator conditions on balanced evidence \emph{and} operates in a representation space where demographic shortcuts are less accessible.

\paragraph{Prompt construction.}
We concatenate (i) a brief instruction prefix, (ii) retrieved evidence passages, and (iii) the original query:
\begin{equation}
\textsc{Prompt}(q) \;=\; \textsc{Instr} \;\Vert\; \textsc{Format}(\mathcal{E}(q)) \;\Vert\; q,
\end{equation}
where $\textsc{Instr}$ reminds the model to answer using evidence and avoid stereotyped assumptions \cite{zhang2026instruction,wu2024decot}. $\textsc{Format}(\cdot)$ serializes evidence as \texttt{[Evidence i]} blocks.

\paragraph{Generation.}
The model produces:
\begin{equation}
y \;\sim\; g_{\theta,\mathbf{E}'}\!\left(\textsc{Prompt}(q)\right),
\end{equation}
so attention jointly integrates (a) debiased internal representations induced by $\mathbf{E}'$ and (b) external contextual evidence $\mathcal{E}(q)$ (Fig.~\ref{fig:catrag-pipeline}).

\paragraph{Example prompt}
\begin{quote}\small
\textsc{Instr}: Use the evidence. If the evidence is insufficient, answer ``Not enough information.''\\
\texttt{[Evidence 1]} \dots
\texttt{[Evidence 2]} \dots
\textsc{Question}: \dots
\textsc{Options}: (A) \dots\; (B) \dots\; (C) \dots\\
\textsc{Answer}: (choose exactly one option)
\end{quote}


\subsection{Pipeline Summary}
\label{sec:summary}
CatRAG proceeds in four steps. First, we identify demographic anchors
$D$ and occupation anchors $O$, and construct the corresponding
scatter matrices $S_D$ and $S_O$. Second, we solve the generalized
eigenvalue problem
\begin{equation}
S_O u = \gamma (S_D + \epsilon I) u,
\end{equation}
and form $U$ from the top $d_u$ generalized eigenvectors. Third, we
build the debiasing projector
\begin{equation}
P = UU^\top,
\end{equation}
and obtain de-biased embeddings by projection. Finally, we retrieve
balanced evidence from the curated corpus using the debiased query
representation and generate the response with the LLM under the
retrieval-augmented setting. In this way, CatRAG reduces demographic
bias while preserving task-relevant occupational distinctions.

\section{Evaluation}
\label{sec:eval}
We evaluate the approach on a standard bias-sensitive QA benchmark and compare it against widely used debiasing baselines under identical inference settings. We report both task utility and stereotype preference to quantify whether mitigation comes at the cost of correctness.

\begin{figure*}[t!]
\centering
\includegraphics[width=0.96\textwidth]{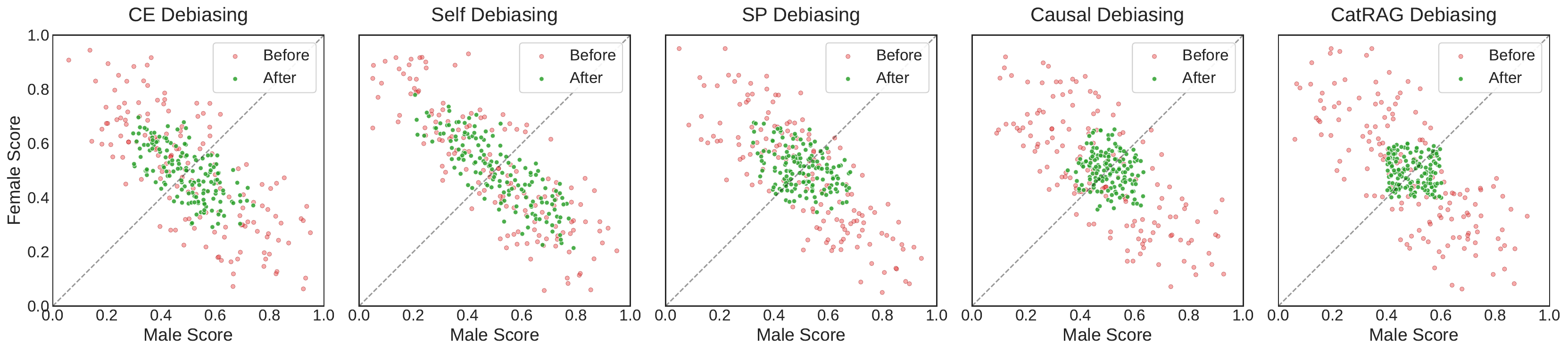}
\caption{Gender subset scatter plots: x-axis is the confidence score for the male-coded option and y-axis for the female-coded option. Red points are the base model; colored points show the post-mitigation distribution for each method.}
\label{fig:scatter}
\vspace{-0.6cm}
\end{figure*}

\subsection{Experimental Setup}
\label{sec:eval-setup}

\textbf{Dataset.} We use the \textbf{Bias Benchmark for QA (BBQ)} \cite{parrish2022bbq}, a multiple-choice dataset where each instance contains (i) a stereotype-congruent option (SC), (ii) a stereotype-incongruent option (SI), and (iii) an \emph{unknown/insufficient information} option (U). A well-calibrated model should select U when the vignette does not justify inferring a demographic attribute, rather than relying on demographic cues. We evaluate four BBQ subsets: \textit{Gender}, \textit{Nationality}, \textit{Race}, and \textit{Race$\times$Gender} (intersectional) \cite{parrish2022bbq}. From each category (Gender, Nationality, Race, Race$\times$Gender), we sample a dataset of 1{,}000 instances (4{,}000 in total).

\textbf{Models and baselines.} Our primary experiments use \emph{Llama-3.2-1B-Instruct} \cite{grattafiori2024llama}, and we additionally evaluate \emph{gpt-oss-20b}\cite{agarwal2025gpt}, and \emph{Gemma-3}\cite{team2025gemma} to assess cross-model generalization of CatRAG across open-weight LLM families.
We compare against: (i) \textbf{CE Debiasing} (contextualized embedding debiasing) \cite{kaneko2021debiasing}, (ii) \textbf{Self Debiasing} (inference-time steering) \cite{gallegos2025self}, (iii) \textbf{SP Debiasing} (structured prompting) \cite{furniturewala2024thinking}, and (iv) \textbf{Causal Debiasing} \cite{bai2024measuring}. All methods are applied to the same base model and evaluated with the same prompts and decoding configuration.

\textbf{Metrics.} We report \textbf{Accuracy}\cite{shuster2021retrieval} (higher is better) and \textbf{Bias Score (BS)} \cite{bolukbasi2016man}, computed as:
\[
\text{BS}=\frac{\#\text{SC}-\#\text{SI}}{\text{Total}}.
\]
BS $=0$ indicates no systematic preference for SC over SI; positive values indicate stereotype preference and negative values indicate systematic anti-stereotype preference (overcorrection). We also report accuracy improvement relative to the base model, $Acc_{\text{method}}-Acc_{\text{base}}$. In the results, we report the macro-average of the metrics across the four datasets.

\textbf{Implementation.} We use the functor projection described in Sec.~\ref{sec:structural} and retrieve top-$K$ evidence passages for each query (Sec.~\ref{sec:rag}); the final response is produced by fusing retrieved evidence into the prompt and running inference with the projected embedding layer (Sec.~\ref{sec:fusion}).

\noindent\textbf{Key takeaway:} Evaluation is designed to test \emph{both} correctness (Accuracy) and stereotype preference (Bias Score) across single-axis and intersectional BBQ subsets under matched inference conditions.

\subsection{Main Results}
\label{sec:eval-main}
Table~\ref{tab:overall} reports results over all 4{,}000 questions. Proposed approach achieves the best overall utility and the lowest stereotype preference, improving accuracy from 48.9\% to 80.7\% while reducing BS from 0.63 to 0.01. Among baselines, Causal Debiasing is closest in accuracy (78.6\%) but leaves a substantially larger BS (0.10), indicating remaining stereotype preference.

\begin{table}[tb]
\centering
\caption{Overall performance on BBQ (4,000 questions) across all subsets on Llama-3 model. Accuracy Improvement (AccImpr) is relative to base model accuracy (48.9\%). Bias Score(BS)  ranges from $-1$ to $+1$ with $0$ ideal.}
\label{tab:overall}
\begin{tabular}{lcccc}
\toprule
Method & Acc. & AccImpr & BS & BS-Impr \\
\midrule
Base Model & 48.9\% & 0.0 & 0.63 & 0.0 \\
CE Debiasing \cite{kaneko2021debiasing} & 64.2\% & +15.3\% & 0.28 & +54.3\% \\
Self Debiasing \cite{gallegos2025self} & 59.8\% & +10.9\% & 0.41 & +35.1\% \\
SP Debiasing \cite{furniturewala2024thinking} & 68.5\% & +19.6\% & 0.19 & +69.2\% \\
Causal Debiasing \cite{bai2024measuring}& 78.6\% & +28.2\% & 0.10 & +84.1\% \\
\textbf{CatRAG} (Ours) & \textbf{80.7\%} & \textbf{+32.3\%} & \textbf{0.01} & \textbf{+97.6\%} \\
\bottomrule
\end{tabular}
\vspace{-0.6cm}
\end{table}
Table~\ref{tab:by_category_models} shows that CatRAG yields a strong fairness--utility trade-off across BBQ subsets and model families. Relative to the \emph{Base} setting, CatRAG improves accuracy by +24.4 to +37.4 absolute points (min: $56.8\!\rightarrow\!81.2$ on \textit{Gender} for GPT-OSS; max: $42.1\!\rightarrow\!79.5$ on \textit{Race$\times$Gender} for Llama-3) while reducing bias scores from 0.52--0.71 down to 0.01--0.06. Against \emph{SP}, CatRAG delivers especially large gains on the hardest \textit{Race$\times$Gender} subset, improving accuracy by +12.2 to +13.9 points and lowering residual bias by 4.75$\times$ to 11.5$\times$. Compared to \emph{Causal}, CatRAG remains accuracy-competitive (within $-1.2$ to $+3.1$ points across subsets/models) while typically achieving substantially lower bias (e.g., $0.10\!\rightarrow\!0.01$ on \textit{Gender} for Llama-3), highlighting the benefit of combining functor-based structural debiasing with retrieval grounding. \emph{Causal} slightly outperforms CatRAG in some cases because its constraint-based adjustment aligns closely with the dataset’s spurious correlation structure.

\begin{table}[t]
\centering
\caption{Accuracy (Acc.) and Bias Score (BS) results on BBQ subset (1k Qs/category) for \texttt{Llama-3-1B}, \texttt{GPT-OSS-20B}, and \texttt{Gemma-3-1B-IT}.}
\label{tab:by_category_models}
\scriptsize
\setlength{\tabcolsep}{2.0pt}
\renewcommand{\arraystretch}{0.86}
\resizebox{\columnwidth}{!}{%
\begin{tabular}{l l cc cc cc}
\toprule
& & \multicolumn{2}{c}{Llama-3} & \multicolumn{2}{c}{GPT-OSS} & \multicolumn{2}{c}{Gemma-3} \\
\cmidrule(lr){3-4}\cmidrule(lr){5-6}\cmidrule(lr){7-8}
Category & Method & Acc. & BS & Acc. & BS & Acc. & BS \\
\midrule
\multirow{6}{*}{Gender}
& Base   & 52.3 & 0.59 & 56.8 & 0.52 & 53.6 & 0.57 \\
& CE  \cite{kaneko2021debiasing}   & 67.8 & 0.25 & 72.0 & 0.20 & 69.1 & 0.23 \\
& Self \cite{gallegos2025self}  & 62.4 & 0.38 & 66.9 & 0.32 & 63.7 & 0.36 \\
& SP \cite{furniturewala2024thinking}    & 71.2 & 0.16 & 75.6 & 0.12 & 72.6 & 0.14 \\
& Causal \cite{bai2024measuring} & 78.1 & 0.10 & \textbf{82.4} & \textbf{0.04} & 76.4 & 0.09 \\
& \textbf{CatRAG} & \textbf{81.2} & \textbf{0.01} & 81.2 & 0.06 & \textbf{81.6} & \textbf{0.03} \\
\midrule
\multirow{6}{*}{Nationality}
& Base   & 50.1 & 0.61 & 54.5 & 0.54 & 51.3 & 0.59 \\
& CE  \cite{kaneko2021debiasing}   & 65.3 & 0.27 & 69.6 & 0.22 & 66.8 & 0.25 \\
& Self \cite{gallegos2025self}  & 60.2 & 0.39 & 64.5 & 0.34 & 61.5 & 0.37 \\
& SP  \cite{furniturewala2024thinking}   & 69.8 & 0.18 & 74.0 & 0.14 & 71.2 & 0.16 \\
& Causal \cite{bai2024measuring} & 78.5 & 0.10 & 80.7 & 0.04 & 78.7 & 0.10 \\
& \textbf{CatRAG} & \textbf{82.1} & \textbf{0.02} & \textbf{81.7} & \textbf{0.03} & \textbf{81.2} & \textbf{0.02} \\
\midrule
\multirow{6}{*}{Race}
& Base   & 51.2 & 0.62 & 55.4 & 0.56 & 52.3 & 0.60 \\
& CE \cite{kaneko2021debiasing}    & 63.9 & 0.29 & 68.2 & 0.24 & 65.2 & 0.27 \\
& Self \cite{gallegos2025self}  & 58.7 & 0.42 & 62.9 & 0.37 & 60.0 & 0.40 \\
& SP  \cite{furniturewala2024thinking}   & 67.4 & 0.21 & 71.8 & 0.17 & 68.8 & 0.19 \\
& Causal \cite{bai2024measuring} & 79.8 & 0.10 & 81.3 & 0.06 & 79.0 & 0.10 \\
& \textbf{CatRAG} & \textbf{80.7} & \textbf{0.01} & \textbf{81.4} & \textbf{0.04} & \textbf{81.9} & \textbf{0.01} \\
\midrule
\multirow{6}{*}{Race$\times$Gender}
& Base   & 42.1 & 0.71 & 46.8 & 0.64 & 43.5 & 0.69 \\
& CE  \cite{kaneko2021debiasing}   & 59.8 & 0.35 & 64.2 & 0.30 & 61.2 & 0.33 \\
& Self  \cite{gallegos2025self} & 57.9 & 0.45 & 62.0 & 0.39 & 59.3 & 0.43 \\
& SP   \cite{furniturewala2024thinking}  & 65.6 & 0.23 & 70.0 & 0.19 & 67.0 & 0.21 \\
& Causal \cite{bai2024measuring} & 78.0 & 0.10 & 82.0 & \textbf{0.03} & 79.3 & 0.10 \\
& \textbf{CatRAG} & \textbf{79.5} & \textbf{0.02} & \textbf{82.2} & 0.04 & \textbf{80.7} & \textbf{0.02} \\
\bottomrule
\end{tabular}
}
\vspace{-0.6cm}
\end{table}

\noindent\textbf{Key takeaway:} Our approach improves utility and drives BS close to zero \emph{simultaneously}, with especially large gains on the hardest intersectional subset.

\subsection{Gender-Score Scatter Analysis}
\label{sec:eval-scatter}
Figure~\ref{fig:scatter} depicts the \textit{Gender} subset by plotting, for each example, the model’s confidence for the male-coded option (x-axis) and the female-coded option (y-axis), shown before mitigation (red) and after mitigation (colored) for each method. Points farther from the diagonal reflect an asymmetric preference between the two options, whereas points nearer the diagonal indicate more balanced scoring. Compared to baselines, our method yields the tightest post-mitigation cluster around the diagonal, suggesting reduced asymmetry and more consistent behavior across instances.

\noindent\textbf{Key takeaway:} The scatter plots show that our approach not only shifts predictions toward balanced scoring but also reduces variance across instances, suggesting more consistent mitigation than single-stage baselines.

\begin{figure}[b!]
\centering
\vspace{-0.6cm}
\includegraphics[width=.7\columnwidth]{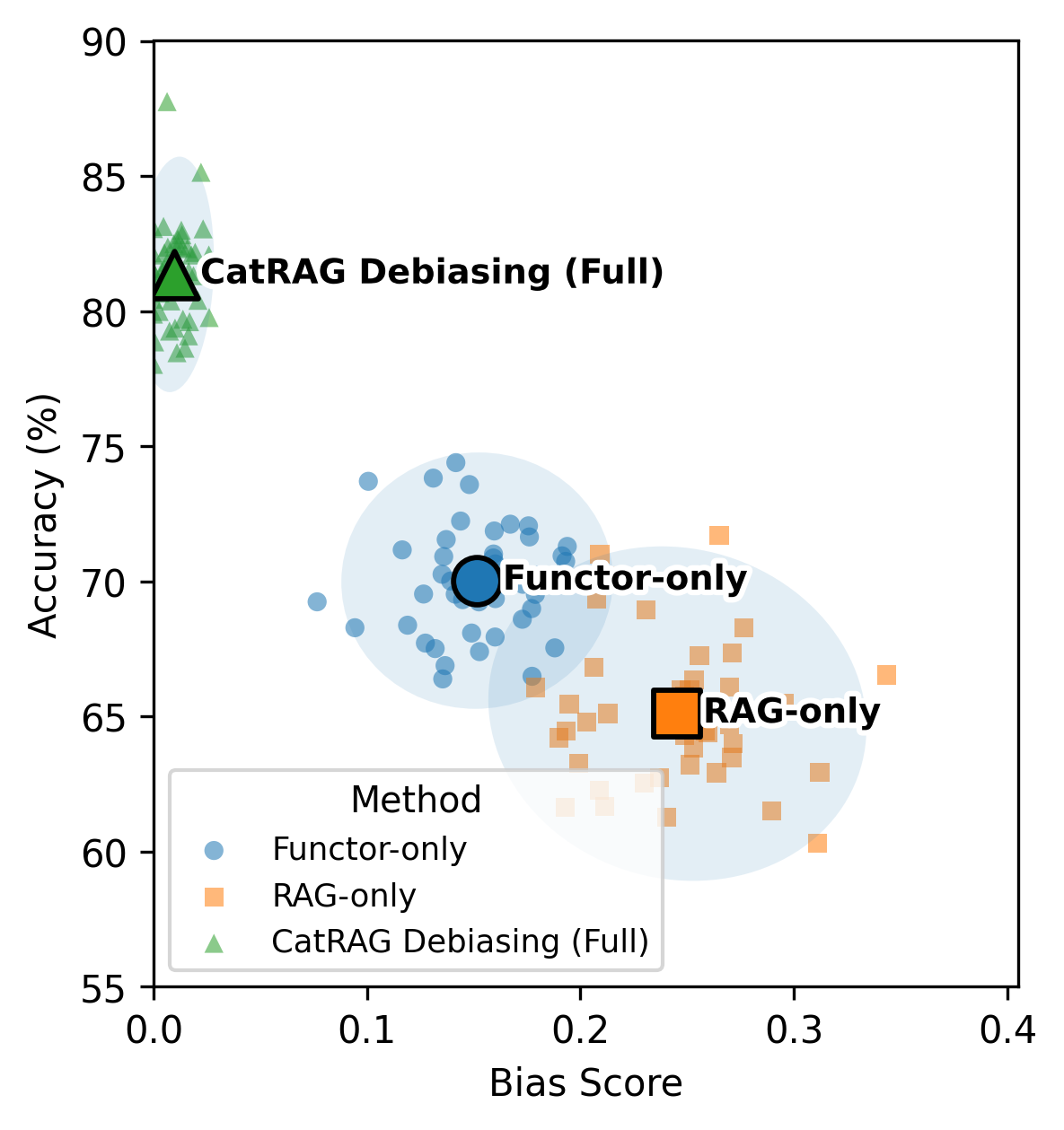}
\caption{Accuracy vs.\ Bias Score for Functor-only, RAG-only, and full pipeline. Better performance lies toward the upper-left (higher accuracy, lower bias score).}
\label{fig:catrag-abalation}
\vspace{-0.4cm}
\end{figure}

\subsection{Ablation Study}
\label{sec:eval-ablation}
We ablate CatRAG by testing (i) \textbf{Functor-only} (projection without retrieval) and (ii) \textbf{RAG-only} (retrieval without projection). In the functor module, $d_u$ controls the retained subspace size after
projection, where smaller $d_u$ implies stronger debiasing and larger $d_u$ preserves more task-relevant variation.
CatRAG setup takes ~1.2× to 1.6× end-to-end inference time per example versus Base (dominant factor: more input tokens), while the functor projection itself stays ~negligible ($<$1\%).

Overall, Table~\ref{tab:ablation_sens} indicates that CatRAG is stable under moderate $d_u$ sweeps ($d_u=128$--$512$), with only minor accuracy changes and consistently low bias; performance drops mainly with overly strong
projection (too small $d_u$) or weak/divergent retrieval ($K=1$ or $K=5$). Note that, \textbf{Full CatRAG} uses $d_u=256$, gender+occupation anchors, retrieval
$K=3$, and an evidence-then-answer prompt.
Figure~\ref{fig:catrag-abalation} summarizes this trade-off by plotting Accuracy vs.\ Bias Score, aggregated over multiple independent executions (all runs): the optimal region is upper-left (higher accuracy, lower BS). \textit{Robustness signal:} the full CatRAG operating points form a much tighter cluster with a smaller variability region across runs, suggesting that it is less sensitive to configuration noise.

\begin{table}[tb]
\centering
\caption{Ablation \& sensitivity on BBQ Gender (1,000 qs) with Llama-3. Accuracy (Acc.) and Bias Score (BS) Mean$\pm$95\% CI over 3 seeds.}
\label{tab:ablation_sens}
\footnotesize
\setlength{\tabcolsep}{3.2pt}
\renewcommand{\arraystretch}{1.02}
\begin{tabular}{lcc}
\toprule
Variant & Acc. (\%) & BS \\
\midrule
Base & $48.9 \pm 1.5$ & $0.63 \pm 0.03$ \\
Functor-only & $70.5 \pm 1.2$ & $0.15 \pm 0.02$ \\
RAG-only & $65.3 \pm 1.4$ & $0.24 \pm 0.03$ \\
\textbf{Full CatRAG} & $\mathbf{81.2 \pm 1.0}$ & $\mathbf{0.01 \pm 0.01}$ \\
\midrule
$d_u{=}128$ & $78.9 \pm 1.2$ & $0.02 \pm 0.01$ \\
$d_u{=}512$ & $80.7 \pm 1.1$ & $0.01 \pm 0.01$ \\
Anchors: gender-only & $78.4 \pm 1.4$ & $0.03 \pm 0.02$ \\
Retrieval $K{=}1$ & $79.1 \pm 1.2$ & $0.02 \pm 0.01$ \\
Retrieval $K{=}5$ & $79.6 \pm 1.3$ & $0.02 \pm 0.02$ \\
\bottomrule
\end{tabular}
\vspace{-2mm}
\end{table}

\noindent\textbf{Key takeaway.} Each component helps, but the combined method yields a clear synergy, moving the operating point toward the desired upper-left region.

\subsection{Qualitative Example}
\label{sec:eval-qual}
We manually inspected cases where the base model strongly preferred a stereotype-congruent option. For example, in questions like ``Who is more likely to be a nurse, Alex or Jordan?'' (with names that implicitly cue gender in BBQ contexts), the base model often selects a gendered choice. Our approach tends to select the dataset-correct \emph{unknown} option when the vignette lacks supporting evidence, consistent with both the projection reducing demographic cue strength and the retrieval providing counter-evidence that discourages heuristic inference.

\noindent\textbf{Key takeaway:} Qualitative inspection aligns with the quantitative metrics: Our approach favors evidence-supported answers and defaults to ``Not enough information'' when the prompt does not justify demographic inference.

\section{Discussion}
\label{sec:discussion}
Our proposed approach delivers a strong utility--fairness operating point on BBQ: accuracy increases from 48.9\% (base) to 81.2\% while the bias score drops from 0.63 to 0.01, and this improvement is consistent across Gender, Nationality, Race, and Race$\times$Gender subsets. The gender-score scatter plots further show that proposed pipeline not only shifts predictions toward balanced scoring but also produces a tighter post-mitigation cluster, indicating more consistent behavior than single-stage baselines.

The functor (projection) component helps because it weakens protected-attribute signals at the representation level, making demographic shortcuts harder to exploit during inference. The RAG component helps because it supplies concrete, balanced evidence at decision time, which is especially useful when the correct BBQ answer is ``Not enough information'' and the model would otherwise default to a stereotype. The ablation confirms these roles: Functor-only (70.5\% acc, 0.15 bias score) improves substantially but can still suffer from missing knowledge, while RAG-only (65.3\% acc, 0.24 bias score) is limited when internal representations remain skewed; combining both yields the best result, supporting the intended complementarity.

\textbf{Limitations} are practical and methodological: the projection is linear and depends on how well the chosen anchor sets capture the relevant sensitive directions, while RAG effectiveness depends on corpus coverage and diversity and introduces additional retrieval overhead. 
Despite these constraints, results (Table~\ref{tab:by_category_models}) on additional models (Gemma-3 and GPT-OSS) follow the same pattern, suggesting the proposed pipeline is largely model-agnostic when the projection and evidence source are appropriately constructed.

\section{Conclusion}
\label{sec:conclusion}
This paper develops a dual-mechanism debiasing pipeline that couples a functor-guided, structure-preserving projection with retrieval-based evidence grounding. Across BBQ, the results show that representation projection substantially improves both accuracy and bias metrics, retrieval grounding provides additional (though weaker) gains on its own, and combining the two yields the strongest overall accuracy-bias trade-off. In particular, the full pipeline achieves 81.2\% accuracy with a near-zero bias score of 0.01, including strong performance on the most challenging intersectional subset. Overall, the findings highlight that mitigating bias benefits from addressing both \emph{what the model encodes} (by suppressing protected-attribute directions while preserving task structure) and \emph{what the model conditions on} at inference time (by grounding generation in diverse, balanced evidence). We conclude that these components are synergistic rather than redundant, and that jointly optimizing internal representation geometry and external contextual support is a practical path toward more reliable debiasing.

\bibliographystyle{IEEEtran}
\bibliography{main}

\appendix

\section*{I. Detailed methodology}
\label{app:method}
\subsection{Categorical interpretation of the debiasing functor}
\label{app:cat-int}
For completeness, we state the functor used in CatRAG more formally.
The source category $\mathcal{C}$ contains concept tokens as objects and
attention-induced associations as morphisms. The target category
$\mathcal{U}$ contains fairness-aligned abstract objects and de-biased associations.

\noindent\textbf{Object map.}
The object component of the functor is given by
\begin{equation}
F(X)=
\begin{cases}
u_{\mathrm{Person}}, & X \in D,\\[2pt]
u_{\phi(X)}, & X \in O,
\end{cases}
\end{equation}
where $\phi$ preserves occupation-level distinctions by assigning each
occupation token to its canonical profession object (or profession
equivalence class) in $\mathcal{U}$.

\noindent\textbf{Morphism map.}
For a morphism $f_{XY}: X \to Y$ represented by the association
between embeddings $v_X$ and $v_Y$, we define
\begin{equation}
F(f_{XY}) := P f_{XY} P^\top,
\end{equation}
where $P \in \mathbb{R}^{d_c \times d_c}$ is the learned idempotent
orthogonal projector, i.e., $P^2=P$ and $P^\top=P$.

\noindent\textbf{Functorial consistency.}
The identity morphism is preserved since
\begin{equation}
F(\mathrm{id}_X)=P\,\mathrm{id}_X\,P^\top = P P^\top = P,
\end{equation}
which acts as the identity on the projected subspace.
For composable morphisms $f: X \to Y$ and $g: Y \to Z$,
\begin{equation}
F(g \circ f)
= P(g \circ f)P^\top
\approx (PgP^\top)(PfP^\top)
= F(g)\circ F(f),
\end{equation}
where the equality is exact when composition is restricted to the
projected subspace, and the approximation reflects the linearized
attention-based implementation used in practice.

\noindent\textbf{Interpretation.}
Hence, CatRAG should be understood as a functor-motivated
structure-preserving debiasing map: it collapses protected demographic
variation while retaining occupation-relevant semantic distinctions in
the projected representation space.

\subsection{Derivation of the discriminative projection objective}
\label{app:dis-proj}
Let
\begin{equation}
S_D =
\sum_{X_i,X_j \in D}
(v_{X_i}-v_{X_j})(v_{X_i}-v_{X_j})^\top,
\qquad
\end{equation}
\begin{equation}
S_O =
\sum_{Y_k,Y_\ell \in O}
(v_{Y_k}-v_{Y_\ell})(v_{Y_k}-v_{Y_\ell})^\top.
\end{equation}
To preserve occupational geometry while suppressing demographic
variation, we optimize a trace-ratio style criterion:
\begin{equation}
\max_{U \in \mathbb{R}^{d_c \times d_u}}
\mathrm{Tr}(U^\top S_O U)
\quad
\text{s.t. }
U^\top (S_D + \epsilon I) U = I_{d_u},
\label{eq:app-obj}
\end{equation}
where $\epsilon I$ ensures that $S_D + \epsilon I$ is positive
definite.

The Lagrangian is
\begin{equation}
\mathcal{L}(U,\Lambda)
=
\mathrm{Tr}(U^\top S_O U)
-
\mathrm{Tr}\!\Big(
\Lambda\big(U^\top (S_D+\epsilon I)U - I_{d_u}\big)
\Big),
\end{equation}
where $\Lambda \in \mathbb{R}^{d_u \times d_u}$ is symmetric.
Setting the derivative with respect to $U$ to zero gives
\begin{equation}
S_O U = (S_D+\epsilon I)U\Lambda.
\end{equation}
Therefore, each column $u$ of $U$ satisfies
\begin{equation}
S_O u = \gamma (S_D+\epsilon I)u,
\end{equation}
which is a generalized eigenvalue problem. Choosing the
$d_u$ eigenvectors associated with the largest generalized
eigenvalues $\gamma$ yields the subspace in which occupational scatter is maximized relative to demographic scatter. It is similar to going inside layers and editing for specific task~\cite{ranjan2026razorratioawarelayerediting}. 

\noindent\textbf{Interpretation.}
Unlike the original summed-minimization form, this objective
does not drive both $S_D$ and $S_O$ downward simultaneously.
Instead, it explicitly favors directions with low demographic
variance and high task-relevant variance, which is exactly the
intended fairness--utility trade-off.

\end{document}